\def\year{2021}\relax
\documentclass[letterpaper]{article} 
\usepackage{aaai21}  
\usepackage{times}  
\usepackage{helvet} 
\usepackage{courier}  
\usepackage[hyphens]{url}  
\usepackage{graphicx} 
\usepackage{natbib}  
\usepackage{caption} 
\usepackage{xspace}

\newif\ifjournal

\journaltrue

\title{Strategy Proof Mechanisms for Facility Location\\in Euclidean and Manhattan Space}
\author {
\ifjournal
Toby Walsh
\else
Anonymous
\fi
\\
}

\affiliations{
\ifjournal
UNSW Sydney and Data61\\
tw@cse.unsw.edu.au
\else
    Affiliation \\
    Address \\
    anonymous@email.com
\fi
}

\newtheorem{mytheorem}{Theorem}
\newtheorem{mylemma}{Lemma}

\newcommand{\myproof}{\noindent {\bf Proof:\ \ }}
\newcommand{\mymax}{\mbox{\rm max}}

\newcommand{\myqed}{\mbox{$\diamond$}}
\newcommand{\myOmit}[1]{}

\newcommand{\innerpoint}{\mbox{\sc InnerPoint}\xspace}

\newcommand{\percentile}{\mbox{\sc Percentile}\xspace}
\newcommand{\myendpoint}{\mbox{\sc EndPoint}\xspace}
\newcommand{\median}{\mbox{\sc Median}\xspace}
\newcommand{\leftmost}{\mbox{\sc Leftmost}\xspace}
\newcommand{\rightmost}{\mbox{\sc Rightmost}\xspace}

\newcommand{\serialdictator}{\mbox{\sc SD}\xspace}

\newcommand{\proportional}{\mbox{\sc Proportional}\xspace}
\newcommand{\invprop}{\mbox{\sc InverselyProportional}\xspace}
\newcommand{\onecentre}{\mbox{\sc OneCentre}\xspace}

\begin{document}

\maketitle

\begin{abstract}
We study the impact on mechanisms for facility location of moving from one dimension to two (or
more) dimensions and Euclidean or Manhattan distances. We consider
three fundamental axiomatic properties: anonymity which is a basic 
fairness property, Pareto optimality which is one of the most important
efficiency properties, and strategy proofness which ensures agents
do not have an incentive to mis-report.  We also consider how well such mechanisms
can approximate the optimal welfare. Our results are somewhat 
negative. Moving from one dimension to two (or more) dimensions
often makes these axiomatic properties more difficult to achieve. 
For example, with two facilities in Euclidean space
or with just a single facility in Manhattan space, 
no mechanism is anonymous, Pareto optimal
and strategy proof. By contrast, mechanisms on the line
exist with all three properties. We also show that approximation ratios may increase
when moving to two (or more) dimensions. 
All our impossibility results are {\em minimal}. 
If we drop one of the three axioms (anonymity, Pareto optimality
or strategy proofness)  {\em multiple}
mechanisms satisfy the other two axioms. 
\end{abstract}

\section{Introduction}

The facility location problem captures many real world problems such as locating
hospitals, telephone exchanges, ambulances, post offices
or playgrounds. 
Beyond geographical domains, there are a range of other real world
problems modeled by facility location problems such as choosing
tax-rates, room temperatures, or members of a committee. 
The facility location problem has attracted researchers from a range of areas including AI, operations
research and social choice (e.g. \cite{flp, egktps2011, ptacmtec2013,fsaaai2015,gnpijcai18}).
Our goal here is to design strategy proof mechanisms that 
elicit the true locations of a set of agents and use this 
information to locate one or more facilities 
to serve the agents  fairly and efficiently.  
In particular, we look to minimize the total or maximum distance
of the agents from the facility serving them. 
These are respectively an utilitarian or egalitarian welfare objective. 

In previous work on mechanism design for facility location problems,
researchers have often limited their attention to the
one dimensional problem, locating facilities on a line (e.g. \cite{ptacmtec2013}).  This captures a
number of real world settings
such as locating distribution centres along a motorway,
or sewage plants along a river. However, many real world
problems have two (or more) dimensions, and use 
metrics such as Euclidean or Manhattan distances. This focus on
one dimensional problems has been justified as a starting point to
consider more complex settings (e.g., trees and networks), and 
as it may provide insight into these more complex
settings (e.g. lower bounds for 1-d problem are inherited
by 2-d problem). Our contribution here is to suggest
caution in such arguments. We show, for example, that 
many Pareto optimal and strategy proof mechanisms do not
lift from 1-d to 2-d space. The
extra freedom provided by an additional dimension 
can make it harder to achieve desirable axiomatic
properties like Pareto optimality and strategy proofness. 

\section{Formal Background}

We have $n$ agents located in 2-d space, and
want to locate $m$ facilities to serve all $n$
agents. Agent $i$ is at location $(x_i,y_i)$.
Distances are either Euclidean or Manhattan:
$d((x,y),(u,v)) = \sqrt{(x-u)^2+(y-v)^2}$ or
$d((x,y),(u,v)) = |x-u|+|y-v|$. 
The extension of the problem to higher dimensions is 
straightforward, while the restriction to
1-d space simply requires setting all y-coordinates to zero. 
In this case, we suppose agents are ordered so that $x_1 \leq \ldots
\leq x_n$. 
An agent is served by the
nearest facility, and a solution is a location $(u_j,v_j)$ for each
facility $j$. 
We let $a_i \in  [1,m]$ be the facility serving agent $i$.
We consider an utilitarian welfare objective of the total distance,
$\sum_{i=1}^n d((x_i,y_i),(u_{a_i},v_{a_i}))$ and an egalitarian
welfare objective of the maximum distance, $\mymax_{i=1}^n 
d((x_i,y_i),(u_{a_i},v_{a_i}))$. The goal is to locate facilities to 
minimize one of these two distances. 

We consider a number of mechanisms for the facility location
problem. 
Given a fixed order over the agents, the \serialdictator\ mechanism is
a serial dictatorship that allocates
the first facility to the location of the first agent, and subsequent
facilities to the next agent in order at a new location. 
In the 1-d problem, 
the \percentile\ mechanism is a family of mechanisms 
with parameters $p_1$ to $p_m$ that
locate facility $j$ at $x_{1+\lfloor p_j (n-1) \rfloor}$ for $j  \in [1,m]$. 
For instance, with $m=1$, the \leftmost\ mechanism has $p_1=0$,
the \median\ mechanism has $p_1=\frac{1}{2}$, and the \rightmost\ mechanism has $p_1=1$.
And, with $m=2$, the \myendpoint\ mechanism, which locates facilities at the left and
rightmost agents, has $p_1=0$ and $p_2=1$.
The $d$-dimensional \percentile\ mechanism picks
an orthogonal set of $d$ axes and applies a \percentile\ mechanism to each
axis. 
For example, the 2-dimensional \median\ mechanism picks an
orthogonal set of axes, and applies the \median\ mechanism to each axis.

In many facility location problems, facilities may also have capacity
constraints (e.g. \cite{capacitated2001,aclpwine2019, acllwaaai2020}). 
A telephone exchange can connect at most a fixed number of houses, 
a kindergarten has places for only a given number of children,
a doctor's practice can serve just a limited number of patients, 
etc. Therefore, we also consider 
an extension of the facility location problem that
includes capacity constraints. This extension was previously
studied in \cite{aclpwine2019,acllwaaai2020}.  
In this capacitated setting, the $j$th facility can serve up to $c_j$
agents. We assume that $n \leq  \sum_{j=1}^m c_i$ so that every agent
can be served.
One special setting we consider is when there is no
spare capacity (i.e. $\sum_{j=1}^m c_j = n$). Another special setting
we consider is when facilities are identical (i.e. $c_i=c_j$ for all
$i<j)$. 
A solution in the capacitated setting is now both a location $(u_j,v_j)$ for each
facility $j$, and an assignment of agents to facilities such that
the capacity constraint on each facility is not exceeded. Agents
no longer have to be served by their nearest facility. 
Let 
$N_j$ denote the set of agents assigned to facility $j$, i.e.,
$N_j = \{i | a_i = j\}$. Then the capacity constraints ensure $|N_j| \leq c_j$ for all $j \in [1,m]$. 
The goal in the capacitated setting is to locate facilities and assign
agents to facilities to minimize either the total or maximum distance.

We consider three desirable axiomatic properties of mechanisms for
facility location:
anonymity, strategy proofness and Pareto optimality. 
Anonymity is a simple but fundamental fairness property that requires 
all agents to be treated the same. 
Pareto optimality is a simple efficiency property that is one of the most fundamental normative properties
in the whole of economics. It demands that we cannot make 
one agent better off without making other agents worse off. 
Finally, strategy proofness is a fundamental game theoretic property
that ensures there is no incentive for agents to act strategically by
mis-reporting their location. 

More formally,
a mechanism is {\em anonymous} iff permuting the agents does not
change the solution. 
It is not hard to see that the \serialdictator\ mechanism is not
anonymous, but that any \percentile\ mechanism is.
A mechanism is {\em Pareto optimal} iff 
it returns solutions that are always Pareto optimal. 
A solution is {\em Pareto optimal} iff there is no
other solution in which one agent travels a strictly shorter
distance, and every other agent travels the same or shorter distance to the facility
serving them.  For example,  the \serialdictator\ mechanism is 
Pareto optimal. 
 A mechanism is {\em strategy proof} 
iff agents cannot improve the solution by
mis-reporting
their location (i.e. no agent can mis-report and reduce the distance to travel to
the facility serving them). 
For example, the $d$-dimensional \percentile\ mechanism is 
strategy proof with any parameters (Theorem 3 in \cite{percentile}). 

Finally, we will consider how well a mechanism approximates the
optimal possible welfare. A mechanism achieves an approximation ratio $\rho$ for the total (maximum)
distance iff the total (maximum) distance in any solution it returns
is at most $\rho$ times the optimal. In this case, we say that the mechanism
$\rho$-approximates the optimal total (maximum) distance. 
For example, in 1-d space, any \percentile\ mechanism such as the
\leftmost\  mechanism 2-approximates
the optimal maximum distance \cite{ptacmtec2013}.

\section{One Facility in Euclidean Space}

We begin with one of the simplest possible settings: 
a single uncapacitated facility in Euclidean space. This is a
setting in which several insightful results are already known
about strategy proof mechanisms.

\subsection{Anonymity and Pareto Optimality}

With an odd number of agents, a mechanism for 
locating a single facility in 2-dimensional Euclidean space is 
anonymous, Pareto optimal and strategy proof iff it is 
a 2-dimensional \median\ mechanism \cite{twodflp}.  For two dimensions and an
even number of agents, or for three or more dimensions and an odd or
even number of agents, no mechanism is 
anonymous, Pareto optimal and strategy proof \cite{twodflp}. 

We can consider dropping in turn anonymity, Pareto optimality and strategy
proofness. 
If we drop anonymity, there are {\em multiple} mechanisms for
Euclidean space that
are Pareto optimal and strategy proof. For instance, 
any \serialdictator\ mechanism is Pareto optimal and strategy
proof but not anonymous. 
Similarly, if we drop Pareto optimality, 
there are {\em multiple} mechanisms for Euclidean space that
are anonymous and strategy proof but not Pareto optimal. For instance, 
the 2-dimensional \leftmost\ and the 2-dimensional \rightmost\ mechanisms
are both anonymous and strategy proof, but not Pareto optimal. 
Finally, if we drop strategy proofness, 
there are {\em multiple} mechanisms for Euclidean space that
are anonymous and Pareto optimal but not strategy proof. For instance, 
consider a mechanism that locates the facility at 
the location of the agent with smallest $x$ coordinate, 
tie-breaking by the smallest $y$ coordinate. This mechanism is 
anonymous and Pareto optimal but not strategy proof. 
We conclude therefore that, with an odd number of agents in Euclidean space, 
anonymity, Pareto optimality and strategy
proofness are a {\em minimal} combination of axioms  
characterizing the 2-dimensional \median\ mechanism. 

\subsection{Total Distance}

With a single facility in Euclidean space, the total distance is minimized
by locating the facility at the geometric median. 
A mechanism locating a single facility 
at the geometric median is anonymous and Pareto
optimal, but not strategy proof \cite{ghc2020}. 
Consider, for instance, agents at $(0,0)$, $(0,2)$, $(12,0)$ and
$(12,2)$. The agent at $(12,0)$ has an incentive
to mis-report their location as $(12,2)$ to move the
geometric median from $(6,1)$ to $(12,2)$ which is nearer to $(12,0)$. 
Thus, a strategy proof mechanism in Euclidean space cannot
always locate a single facility so as to minimize the total distance.
We contrast this with the 1-d case where the \median\ mechanism
is strategy proof and returns the optimal total distance. 

As in \cite{ptacmtec2013}, 
we might ask: since a strategy proof mechanism in Euclidean space
cannot be optimal with respect to the total distance, how well can it approximate
the optimal total distance? 
We prove next that no strategy proof
mechanism can be better than an
$\alpha$-approximation for some 
$\alpha$ strictly greater than 1. 

\begin{mytheorem}
In Euclidean space, 
there exists an $\alpha > 1$ such that no deterministic and strategy
proof mechanism for
locating a single facility is an $\alpha$-approximation or better of the optimal total
distance. 
\end{mytheorem}
\myproof
\ifjournal
Assume the opposite. For any $\alpha > 1$, 
there exists a deterministic and strategy proof mechanism that is an $\alpha$-approximation
of the optimal total distance. 
Consider four agents at $(0,0)$, $(0,2)$, $(12,0)$ and $(12,2)$. 
The optimal placement of the facility is at the geometric median: $(6,1)$.
The total distance of the four agents from the geometric median is 
$4\sqrt{37}$. 
Suppose the facility moves one unit distance from the
geometric median. Then the total distances of the agents from the facility increases to
at least $2(\sqrt{50}+\sqrt{26})$. This smallest total distance occurs when the
facility is at $(5,1)$ or $(7,1)$.  
The ratio between these two total distances
is $\frac{\sqrt{50}+\sqrt{26}}{2\sqrt{37}}$. 
This is strictly between $1.0003$ and $1.0004$. 
Set $\alpha=1.0003$. Then the mechanism 
must locate the facility within one unit of distance from the 
geometric median, and a distance of over five units from $(12,0)$. 

Suppose now the agent at $(12,0)$ mis-reports their location as
$(12,2)$. The geometric median moves to $(12,2)$. 
The total distance of the reported positions of the 
four agents from $(12,2)$ is
$12 + 2\sqrt{37}$ which is approximately $24.166$. 
If the mechanism locates
the facility more than two units distance from $(12,2)$
then algebraic reasoning shows that the total distance of the facility
from the reported positions of the agents increases from 
$12 + 2\sqrt{37}$ to more than 24.18. 
The ratio between these total distances is greater than 1.0005. 
Hence, if the mechanism is an $\alpha$-approximation
for $\alpha = 1.0003$, 
then the facility must be located less than two units distance from
$(12,2)$. But the facility in this case is strictly less than four units 
distance from $(12,0)$. Hence the agent at $(12,0)$ has an incentive
to mis-report their location as $(12,2)$. This contradicts the assumption that
the mechanism is strategy proof. Therefore the initial assumption that
it is an $\alpha$-approximation for $\alpha=1.0003$ must be false. 
\fi
\myqed

This result shows that there is a lower bound on the approximation
ratio of any deterministic and strategy proof mechanism in Euclidean space. 
It leaves open how large this approximation
ratio must be. For an odd number of agents and 2-dimensional Euclidean
space, the 2-dimensional \median\
mechanism provides an upper bound on the best possible
\ifjournal approximation ratio for the total distance. In particular, 
\fi
the 2-dimensional \median\
mechanism returns an $\frac{\sqrt{2}\sqrt{n^2+1}}{n+1}$-approximation of the optimal
total distance 
\cite{ghc2020}. By letting $n$ tend to infinity, we conclude that $\alpha \leq \sqrt{2}$. 
We contrast this with the 1-d setting in which the \median\
mechanism is always strategy proof and optimal, locating the facility
to minimize the total distance.

\subsection{Maximum Distance}

In Euclidean space, bounds on the performance of strategy proof
mechanisms in computing the optimal maximum distance are less
well studied.
It is again easy to see that no strategy proof mechanism can return the optimal solution. 
For instance, the \onecentre\ mechanism, 
which locates the facility at the centre of the smallest enclosing
circle, returns an optimal solution for agents on the Euclidean plane. 
The smallest enclosing circle can be found in linear time \cite{onecentre}. 
While this mechanism
is anonymous and Pareto optimal, it 
is not strategy proof. Consider, for instance, 
two agents at $(0,0)$ and one at $(0,1)$. The rightmost
agent can mis-report their location as $(0,2)$ to 
achieve a better outcome. In fact, it follows from the 1-d setting
(Theorem 3.2 of \cite{ptacmtec2013}),
that no deterministic
and strategy proof mechanism can do better than 2-approximate the
optimal maximum distance in Euclidean space. 

With an odd number of agents on the Euclidean plane,
the only mechanism that is anonymous, Pareto optimal and strategy proof is
the 2-dimensional \median\ mechanism \cite{twodflp}.  
We now prove that this mechanism
$2$-approximates the optimal maximum distance. 
To show this, we need a lemma that this mechanism 
locates the facility on or within
the smallest enclosing circle. 
To ensure the median is unambiguously defined and the mechanism is
strategy proof, we restrict
ourselves to an odd number of agents.

\begin{mylemma}
With 2-dimensional Euclidean space and an odd number of agents, the 2-dimensional \median\ mechanism 
locates the facility on the circumference of or within 
the smallest enclosing circle. 
\end{mylemma}
\myproof
Suppose that there are $2k+1$ agents,
the facility is at $(x,y)$, 
and the centre of the smallest enclosing
circle is $(x',y')$.
There are two cases. In the first, 
$x \geq x'$. Let the circumference of the smallest enclosing
circle go through $(x,y'+c)$ and
$(x,y'-c)$. There are therefore $k+1$ agents
with $x$ coordinates greater than or
equal to $x$. These $k+1$ agents have a 
$y$ coordinate that is smaller than or equal to $y'+c$
as they are on or within the smallest enclosing
circle. The median $y$ coordinate
of the agents cannot then be larger than $y'+c$.
By a similar argument, 
the median $y$ coordinate
of the agents cannot also be smaller than $y'-c$.
Hence, the facility lies on or within the smallest 
enclosing circle. 
In the second case, $x \leq x'$. 
This case is symmetric to the first. 
\myqed

\begin{mytheorem}
With 2-dimensional Euclidean space and an odd number of agents, the 2-dimensional \median\ mechanism 
$2$-approximates the optimal maximum distance.
\end{mytheorem}
\myproof
By the previous lemma, the facility is located on or within the
smallest enclosing circle. The worst case
is when the facility is located on the circumference
of this circle and there is an agent diametrically
opposite. In this case, this agent is twice the optimal
\ifjournal maximum distance away from the facility. 
\fi
\myqed

It follows that the 2-dimensional \median\ mechanism
is optimal in terms of strategy proof mechanisms for approximating the optimal maximum distance.
No deterministic and strategy proof mechanism has
a better approximation ratio even when restricted to the line
(Theorem 3.2 of \cite{ptacmtec2013}). 
\ifjournal With an odd number of agents and a single facility, moving
from the 1-d to the 2-d setting 
\fi
changes little. The \median\ mechanism on the line
achieves the same (and optimal) 2-approximation 
of the maximum distance as the 2-dimensional \median\ mechanism
does on the Euclidean plane. 

\section{Multiple Facilities in Euclidean Space}

We next consider strategy proof mechanisms for locating two or more uncapacitated facilities in
Euclidean space. There is less
known about this setting than about two or more facilities in one
dimensional Euclidean space,
or about a single facility in an Euclidean space of two or more dimensions. 
\ifjournal Our results here are somewhat negative. \fi
We first prove 
a strong impossibility theorem: 
when locating two or more facilities
in Euclidean space, we cannot simultaneously achieve anonymity,
Pareto optimality and strategy proofness.

\begin{mytheorem}
\label{twoflpEuclidean}
With $m$ uncapacitated facilities ($m \geq 2$), $n$ agents ($n \geq m+3$) and Euclidean space with
two or more dimensions, 
no mechanism for facility location is anonymous, Pareto optimal and
strategy proof. 
\end{mytheorem}
\myproof
We consider just two dimensional Euclidean space. The proof for
three or more dimensions merely needs to restrict agents
to a plane. 
Suppose we have 2 facilities and $n$ agents with $n \geq 5$. 
Either $n=2k+2$ or $n=2k+1$. 
Put $2k$ agents at $(0,0)$ and the remaining $n-2k$ agents (which is
one or two at most) at $(100,100)$. 
The unique Pareto optimal mechanism locates the facilities
at $(0,0)$ and $(100,100)$ serving the agents at each location. 
Suppose the agents at $(100,100)$ are
fixed but those at $(0,0)$ are allowed to move
in the box bounded between $(0,0)$
and $(1,1)$. Then essentially we have a single facility location
problem on the remaining $2k$ agents that locates
the leftmost facility somewhere in the convex hull of the 
reported locations of the $2k$ agents. 
Theorem 4.1 in \cite{twodflp} demonstrates that 
no mechanism is anonymous, Pareto optimal and strategy proof 
for an even number of agents in two dimensional Euclidean 
space. This proof continues to hold if the locations
of the agents are limited
to a small box such as between $(0,0)$ and 
$(1,1)$. Note that an agent cannot profitably
mis-report their location as $(100,100)$. To ensure anonymity
and Pareto optimality, the mechanism must then locate both
facilities at $(100,100)$ which puts the nearest
facility surely at a greater distance. A sincere report
puts the facility within the convex hull of the leftmost
$2k$ agents which is nearer to their location than
$(100,100)$. 
Hence, there is no mechanism 
that is anonymous, Pareto optimal and strategy proof for two
facilities in the Euclidean plane. 
For $m$ facilities with $m>2$, we place
an additional agent at $(100j,100j)$ for $j=3$ to $m$. 
\myqed

We contrast this impossibility result with the 1-d setting where anonymity, Pareto
optimality, strategy proofness and good approximation ratios are 
all simultaneously achievable. 
In particular,  when locating two facilities on the line, the \myendpoint\ mechanism 
is anonymous, Pareto optimal and strategy proof, and has 
good approximation ratios for the total and maximum distances \cite{ptacmtec2013}. 
It is somewhat disappointing then that, when we
move from two facilities in 1-d Euclidean space to two facilities in 2-d Euclidean space, 
we can no longer simultaneously achieve these desirable axiomatic properties. 

We also contrast this impossibility result with the setting of a single facility in
the Euclidean plane where anonymity, Pareto
optimality, strategy proofness and good approximation ratios are 
again all simultaneously achievable provided we have an odd number of
agents. 
In particular, the two-dimensional \median\ mechanism 
satisfies all these desirable axiomatic properties. 
Note that Theorem \ref{twoflpEuclidean} continues to hold 
in the Euclidean plane with just an odd (or just an even) number of
agents.  It is again somewhat disappointing then that, when we
move from one facility in the Euclidean plane to two facilities in the
Euclidean plane, we can no longer simultaneously achieve these desirable axiomatic properties. 

We next consider dropping in turn anonymity, Pareto optimality
and strategy proofness. 
If we drop anonymity, there are {\em multiple} 
mechanisms for locating two
or more facilities in Euclidean space that are Pareto optimal and
strategy proof but not anonymous. For example, any \serialdictator\ mechanism is 
Pareto optimal and strategy proof but not anonymous. 
Similarly, if we drop Pareto optimality, there are {\em multiple}
mechanisms  that are anonymous and strategy proof but not Pareto
optimal (e.g. any multi-dimensional \percentile\
mechanism with parameters $p_i=p_j$ for all $i$ and $j$). 
And if we drop strategy proofness, there are {\em multiple} mechanisms 
that are anonymous and Pareto optimal but not strategy proof. 
Consider, for example, the mechanism
that picks some orthogonal set of coordinate axes,
then orders agents by $x$-coordinate, tie-breaking by $y$-coordinate,
and locates the first facility at the position of the first agent in this order, and
subsequent facilities at the position of the next agent in the order
at a new location. We therefore conclude that anonymity, Pareto optimality and strategy
proofness are a {\em minimal} combination of incompatible axioms
for locating two or more facilities in 
Euclidean space.

\section{Welfare Bounds}

We now consider how well strategy proof mechanisms can approximate the 
optimal welfare when locating multiple facilities
in Euclidean space. For instance, can we bound the approximation ratio
for the maximum or total distance?

When locating two facilities on the line, the only deterministic, anonymous and strategy proof mechanism
with a bounded approximation ratio for the total distance
is the \myendpoint mechanism (Theorem 3.1 in \cite{ft2013}).
Recall that the \myendpoint\ mechanism is an instance of the
\percentile\ mechanism with $m=2$, $p_1=0$ and $p_2=1$. 
The \myendpoint\ mechanism also provides a 2-approximation of
the optimal maximum distance. Again, this is optimal as no deterministic and strategy proof
mechanism can have a better approximation ratio (Corollary 4.4 of
\cite{ptacmtec2013}). In fact, the \myendpoint\ mechanism is not just
optimal in terms of the approximation ratio of the maximum distance. It is the only
\percentile\ mechanism for 
locating two uncapacitated facilities on the line with a {\em bounded} 
approximation ratio for the maximum distance \cite{anonflp}.

We therefore consider next welfare bounds for 
multi-dimensional \percentile\ mechanism when locating two or more facilities
in Euclidean space. 
It is known that in Euclidean space the multi-dimensional \percentile\ mechanism for two or
more uncapacitated
facilities has an unbounded approximation ratio 
for the total distance (Theorem 4 in \cite{percentile}). 
Can we do better in approximating the maximum distance?
This is certainly the case in 1-dimension where we can approximate
the maximum distance better than the total distance. Specifically, 
with any mechanism on the line that is deterministic and strategy proof, the optimal 
approximation ratio for the total distance 
is $n-2$, but for
the maximum distance is just 2 \cite{ft2013}. We prove next that, in 
Euclidean space, multi-dimensional
\percentile\ mechanisms cannot do better at approximating
the maximum distance than they can at approximating the 
total distance. 

\begin{mytheorem} \label{identity}
No multi-dimensional \percentile\ mechanism for two (or more)
facilities in two (or more) dimensional 
Euclidean space bounds the approximation ratio for the maximum distance. 
\end{mytheorem}
\myproof
The maximum distance, $d_{max}$ is lower bounded by the following
identity relating it to $d_{total}$, the total distance: 
$d_{max} \geq \frac{d_{total}}{n}$. Hence, if the total distance is
unbounded  (as demonstrated by Theorem 4 in \cite{percentile}, then the maximum distance
is too. Rather than call upon this identity, we will give an explicit proof that the approximation
ratio for the maximum distance is unbounded since
the proof provides
insight into why the approximation ratio can be unbounded, and how
the \percentile\ mechanism can go wrong. Note that each case in this proof is also an example of
how \percentile\ mechanisms do not bound the approximation
ratio for the total distance. 

We consider two facilities in two dimensions. The proof
easily generalizes to more facilities and more dimensions.
There are three cases.
In the first case, neither of the two facilities is at a location corresponding to
parameters $(1,1)$. That is, neither of the two facilities is at 
$(x_{max},y_{max})$ where $x_{max}$ and $y_{max}$ are
the maximum reported $x$ and $y$ coordinates.
Let $p_{max}$ be the largest parameter less than 1,
and $n = \lceil \frac{3}{1-p_{max}} \rceil$. Consider
$n-1$ agents at $(0,0)$ and one at $(1,1)$. 
The optimal maximum distance is zero with
a facility at $(0,0)$ and one at $(1,1)$. 
However, the \percentile\ mechanism returns a solution
with both facilities at $(0,0)$, and 
maximum distance of $\sqrt{2}$. 
In the second case, one of the facilities is at a location
corresponding to  parameters $(1,1)$
and the other is at a location corresponding to $(p_1,p_2)$ where at least one
$p_i$ is non-zero. 
Let $p_{max}$ be the maximum of $p_1$ and $p_2$,
and $n = \lceil \frac{3}{p_{max}} \rceil$. Consider
$n-1$ agents at $(1,1)$ and one at $(0,0)$. 
The optimal maximum distance is zero with
a facility at $(0,0)$ and one at $(1,1)$. 
However, the \percentile\ mechanism returns a solution
with one facility at $(1,1)$, and the other not at $(0,0)$. 
This solution has a maximum distance of $\sqrt{2}$. 
In the third case, 
one of the facilities is at a location corresponding to parameters $(0,0)$
and the other is at a location corresponding to parameters $(1,1)$. Consider 
one agent at $(0,1)$ and the other at $(1,0)$. 
The optimal maximum distance is zero with
a facility at each agent, 
but the \percentile\ mechanism returns a solution
with a maximum distance of 1. 
\myqed

\section{Capacitated Facilities}

We next consider the impact of capacity constraints on facility
location in Euclidean space. 
We might hope to design mechanisms with stronger axiomatic properties
in the capacitated setting than
the uncapacitated setting for at least two reasons.
The first reason is that adding capacity limits requires us to
allocate agents to particular facilities, and this may reduce the opportunity for
agents to be strategic.
For example, in a general metric space, 
the randomized \proportional\ mechanism is not strategy
proof with three or more agents but becomes so when
we make it winner imposing \cite{ft2010}.

A second reason we might do better in the capacitated setting in
Euclidean space is that, with multiple uncapacitated facilities, irrespective
of whether the total number of agents is even or odd, one facility might need to serve
an even number of agents, and no mechanism in Euclidean space is 
anonymous, Pareto optimal and strategy proof with an even number
of agents \cite{twodflp}.  
On the other hand, in the capacitated setting, we can limit ourselves
to situations in which the capacity of each facility is an odd
number. Recall that in the Euclidean plane with an odd number of
agents, there exists an unique mechanism
to locate a single uncapacitated facility
\ifjournal that is anonymous, Pareto optimal and strategy proof 
\cite{twodflp}. 
\fi

Unfortunately, this hope is not realized. 
We prove another strong impossibility
theorem: no mechanism for locating two or more capacitated
facilities in Euclidean space can simultaneously be anonymous,
Pareto optimal and strategy proof even limited to facilities with an odd capacity. 

\begin{mytheorem}
With two or more facilities of capacity 3 or larger and Euclidean space with
two or more dimensions, 
no mechanism for capacitated facility location is anonymous, Pareto
optimal
and strategy proof. 
\end{mytheorem}
\myproof
Suppose that such a mechanism exist. 
We can assume that the facilities have the same size and that
none of their capacity is spare since 
there is already no mechanism that 
is anonymous, Pareto optimal and strategy proof in the one
dimensional setting when either we have two facilities with two or more different sizes,
or we have spare capacity \cite{anonflp}. 
We consider also just two dimensional Euclidean space. The proof for
three or more dimensions merely needs to restrict agents
to a two dimensional plane. 

We suppose that we have two facilities of capacity $c$. 
There are two cases. If $c$ is even, say $c=2k$ with $k>1$, we
consider $2k$ agents in the box between $(0,0)$ and 
$(1,1)$, and another $2k$ agents at $(100,100)$. 
Any anonymous and Pareto optimal mechanism locates one facility  
at $(100,100)$ serving the agents at this location, and the other 
in the box between $(0,0)$ and 
$(1,1)$. Suppose the agents at $(100,100)$ are
fixed. Then essentially we have a single facility location
problem on the remaining $2k$ agents. 
Theorem 4.1 in \cite{twodflp} demonstrates that 
no mechanism is anonymous, Pareto optimal and strategy proof
for an even number of agents in two dimensional Euclidean 
space. This proof continues to hold if the agents are limited
to a small box such as between $(0,0)$ and 
$(1,1)$. 
Note that an agent cannot profitably
mis-report their location as $(100,100)$. To ensure anonymity
and Pareto optimality, the mechanism must then locate both
facilities at $(100,100)$ which puts the 
facility serving this agent at a greater distance. A sincere report
locates the facility within the convex hull of the leftmost
$2k$ agents which is nearer to their true location than
$(100,100)$. 
Hence, there is no mechanism for two facilities
with even capacity in the Euclidean plane
that is anonymous, and Pareto optimal and strategy proof.

If $c$ is odd, say $c=2k+1$ with $k \geq 1$, we 
consider $2k+1$ agents at $(0,0)$, and
another $2k+1$ agents at $(100,100)$. 
Any anonymous and Pareto optimal mechanism locates
one facility at $(0,0)$ and 
the other at $(100,100)$. 
Suppose the agents at $(100,100)$ are fixed. 
Then essentially we have a single facility location
problem on the remaining $2k+1$ agents. 
Theorem 3.1 in \cite{twodflp} demonstrates that 
any mechanism in this setting that is anonymous, Pareto optimal and strategy proof
is a multi-dimensional \median\ mechanism. Whatever orthogonal coordinate 
axes we take, this puts the leftmost facility at $(0,0)$. 
Suppose one agent at $(0,0)$ is moved along the 
diagonal towards the $2k+1$ agents at $(100,100)$. 
The two facilities remain immobile at $(0,0)$ and $(100,100)$. 
Consider now the agent arriving at $(100,100)$. 
By anonymity over the $2k+2$ agents now at $(100,100)$ and Pareto optimality, both 
facilities must be at $(100,100)$. This contradicts
the leftmost facility remaining at $(0,0)$. 
This contradiction
means that our initial assumption, that an anonymous,
Pareto optimal and strategy proof mechanism exists,
is false. 
For $m$ facilities with $m>2$, we consider 
an additional $(m-2)c$ agents at $(200,200)$. 
\myqed

\ifjournal 
An interesting open case is four agents and two identical facilities
with no space capacity. Is there an anonymous, Pareto optimal and strategy proof
mechanism for this setting? 
\fi

We contrast this last impossibility result 
with the 1-d setting. For facility location on the line, anonymity, Pareto optimality and
strategy proofness can
simultaneously be achieved. In particular, with two identical
and capacitated facilities on the line, provided there is no spare capacity, 
the \innerpoint\ mechanism is anonymous, 
Pareto optimal and strategy proof and has good approximation
ratios for the total and maximum distance \cite{acllwaaai2020}. 

Finally, we consider dropping in turn anonymity, Pareto optimality
and strategy proofness. By similar arguments to case of strategy proof
mechanisms for locating multiple uncapacitated facilities in Euclidean space, we can show that
there are {\em multiple} mechanisms satisfying just two of these
axioms. It follows then that anonymity, Pareto optimality and strategy proofness
are a {\em minimal} combination of incompatible axioms for
locating two or more capacitated facilities in two or more dimensional
Euclidean space.

\section{Manhattan distances}

We conclude by switching from Euclidean to Manhattan
distances. We might hope to
achieve stronger axiomatic properties with Manhattan distances 
as these
are somewhat ``closer'' to the one dimensional
setting, and strategy proof mechanisms are somewhat ``easier'' to construct on
the line. Unfortunately, this is not the case. 

For a single facility on the line, 
any mechanism that is anonymous, Pareto optimal and strategy proof 
is a generalized median mechanism
\cite{moulin1980}. 
Barbera {\it et al.} \cite{gmvs} 
lifted this result to Manhattan distances in two or more dimensions. 
In particular, they prove that any strategy proof mechanism
on $m$ dimensions can be decomposed coordinate wise
into strategy proof mechanisms that act on the $m$ coordinates
individually. What if we additionally demand Pareto optimality? 
We prove a strong characterization result: with Manhattan distances and two or more dimensions, 
mechanisms can only be anonymous, strategy proof and Pareto optimal
iff there are three or fewer agents. 

\begin{mytheorem}
For three or fewer agents in two or more dimensions,
there exist anonymous, strategy proof and Pareto
optimal mechanisms for locating a single facility
with Manhattan distances. 
\end{mytheorem}
\myproof
For a single agent, we locate the facility at the agent.

For two agents, consider a mechanism 
which sets each coordinate of the facility to be
the max (or the min) of the two reported coordinates of the two agents.
This mechanism is anonymous and 
strategy proof. To show Pareto optimality, 
there are two cases. Either the facility is 
at the location of one of the agents which is
clearly Pareto optimal, or it is at 
an empty vertex of the bounding hypercube around
the two facilities. Moving along one coordinate direction towards one agent 
reduces the Manhattan distance to this agent, but
increases the Manhattan distance to the other agent. 
Hence, the solution is Pareto optimal. 

For three agents, consider a mechanism 
which sets each coordinate to be the median
of the three reported coordinates of the three agents. 
This mechanism is anonymous and 
strategy proof. To show Pareto optimality, 
there are two cases. 
Either the facility is 
at the location of one of the agents which is
clearly Pareto optimal. Or the facility is in the interior
of the bounding hypercube around
the three facilities. Moving along one coordinate direction towards one agent 
reduces the Manhattan distance to this agent, but
increases the Manhattan distance to at least one of the
two other agents. Hence, the solution is Pareto optimal. 
\myqed

\begin{mytheorem}
For four or more agents in two or more dimensions,
there is no anonymous, strategy proof and Pareto
optimal mechanism for locating a single facility 
with Manhattan distances. 
\end{mytheorem}
\myproof
To show the result for four or more agents in two or more dimensions,
we suppose agents
are limited to the $x$-$y$ plane. 
Suppose we have three agents at 
$(0,2)$, $(1,0)$ and $(2,1)$ respectively. 
Consider a mechanism that locates
a facility at the maximum of each coordinate
axis. This puts the facility at $(2,2)$. This
is Pareto dominated by the solution which
locates the facility at $(1,1)$.
Similarly, the mechanism that locates
a facility at the minimum of each coordinate
axis puts the facility at $(0,0)$. This
is also Pareto dominated by the solution which
locates the facility at $(1,1)$.
If we rotate these three agents 90 degrees,
we can also demonstrate that any mechanism
that puts the facility at the maximum of one 
coordinate axis and the minimum of the other
can return Pareto dominated solutions. 

We now consider a four agent problem made up of 
these three agents and one other. 
By Theorem 2 in \cite{gmvs}, any 
strategy proof mechanism in two dimensions is a coordinate-wise
combination of strategy proof mechanisms. 
By \cite{moulin1980}, it must be a coordinate-wise
combination of generalized median voting schemes. 
We can shift our original three voters in the box 
bounded by $(0,0)$ and $(2,2)$ so that
they are well away from any ``phantom'' voters.
Hence, any strategy proof mechanism for our four agents picks out 
the $j$th ranked $x$-coordinate and the
$k$th ranked $y$-coordinate for some $j, k \in [1,4]$. 
Suppose $j=1$ and $k=2$. Then place
the fourth agent at $(3,0)$. Our strategy proof mechanism
then locates the facility at the minimum $x$-coordinate
and minimum $y$-coordinate of the original three agents. 
That is, it locates the facility at $(0,0)$. 
This is Pareto dominated by the mechanism that
locates the facility at $(1,1)$. Similar constructions
show that any other parameters $j$ and $k$ also return solutions
that are Pareto dominated. Hence, no mechanism for 
four agents in two dimensions 
is anonymous, strategy proof and Pareto
optimal. 
The proof extends to five or more agents by placing
additional agents at appropriate boundary positions. 
\myqed

A similar impossibility result for locating {\em multiple} facilities in 
Manhattan space quickly follows: there is no anonymous,
strategy proof and Pareto optimal mechanism
for locating $m$ uncapacitated facilities ($m \geq 2$) and $m+3$ or more agents
in Manhattan space. 

We next consider dropping in turn anonymity, Pareto optimality
and strategy proofness. 
If we drop anonymity, there are {\em multiple} 
mechanisms for locating one
or more facilities in Manhattan space that are Pareto optimal and
strategy proof but not anonymous. For example, any \serialdictator\ mechanism is 
Pareto optimal and strategy proof but not anonymous. 
Similarly, if we drop Pareto optimality, there are {\em multiple}
mechanisms for Manhattan space that are anonymous and strategy proof but not Pareto
optimal (e.g. any multi-dimensional \percentile\
mechanism with parameters $p_i=0$ for all $i$). 
And if we drop strategy proofness, there are {\em multiple} mechanisms 
that are anonymous and Pareto optimal but not strategy proof. 
Consider again the mechanism
that orders agents by $x$-coordinate, tie-breaking by $y$-coordinate,
and locates the first facility at the position of the first agent in this order, and
subsequent facilities at the position of the next agent in the order
at a new location. We therefore conclude that anonymity, Pareto optimality and strategy
proofness are a {\em minimal} combination of incompatible axioms
for locating one or more facility in two or more dimensional
Manhattan space. 

We end with some discussion about welfare bounds. 
In particular, we consider welfare bounds for 
multi-dimensional \percentile\ mechanism when locating one or more facility
in Manhattan space. 
\ifjournal Recall that such mechanisms are anonymous and
strategy proof. 
\fi
We suppose the mechanism chooses axes that align with those used
to compute Manhattan distances. 
We first observe that lower bounds on approximation ratios for
facility location in Manhattan space can be inherited from lower bounds on approximation
ratios for facility location on the line (e.g. those bounds in \cite{ptacmtec2013}). Hence, 
with a single facility,
no deterministic and strategy proof mechanism 
is better than a 2-approximation for the maximum Manhattan distance. 
In fact, with a single facility, the multi-dimensional \median\ mechanism
is optimal with respect to both the total and the maximum
Manhattan distances. No deterministic and strategy proof mechanism
provides better approximation ratios. 

\begin{mytheorem}
The multi-dimensional 
\median\ mechanism locates a single facility to
minimize the total Manhattan distance, and to 2-approximate the
optimal maximum Manhattan distance. 
\end{mytheorem}
\myproof
With Manhattan distances, the total distance is
minimized by minimizing each coordinate individually.
The \median\ mechanism does this along each coordinate.
For the maximum Manhattan distance, each coordinate at worst
2-approximates its contribution to the overall
\ifjournal Manhattan 
\fi
distance. 
\myqed

When locating more facilities in Manhattan space, approximation becomes more challenging. 
In particular, with two or more uncapacitated facilities in two or more dimensions, 
no multi-dimensional \percentile\ mechanism has a bounded approximation ratio 
for the total Manhattan distance (Theorem 4 in \cite{percentile}). 
It follows quickly that 
no multi-dimensional \percentile\ mechanism 
has a bounded approximation ratio 
for the maximum Manhattan distance
(using the same argument as in the proof of Theorem \ref{identity}). 

We contrast these results with the 1-d setting. With a single
facility, approximation ratios are the same in 1-d as 2-d Manhattan space. 
With two facilities on the line, the \myendpoint\ mechanism has bounded approximation
ratios for the total and maximum distances. When we move to 2-d Manhattan
space, the approximation ratios become unbounded. 
With three or more facilities, approximation ratios are unbounded
whether we are in 1-d or 2-d Manhattan space. 
\ifjournal 
In general, moving from 1-d to 2-d Manhattan space,
tends to increase approximation ratios for the total or maximum
distance. 
\fi
We also contrast results for Manhattan space with those for
Euclidean space. With a single facility, 
the 2-dimensional \median\ mechanism is optimal with respect to the 
total Manhattan distance but is just a $\sqrt{2}$-approximation of the
optimal Euclidean distance. In other settings we observe similar
approximation ratios for Manhattan as Euclidean distances
(e.g. with two or more facilities and multi-dimensional \percentile\
mechanisms, approximation ratios 
are unbounded in both Euclidean and Manhattan space). 

\ifjournal
\section{Other Mechanisms for Multiple Facilities}

A number of specific mechanisms have been proposed
with good axiomatic properties that will locate multiple facilities in
two or higher dimensional space. 
With two facilities and any type of metric space, Lu {\it et al.} prove 
that the randomized \proportional\ mechanism is strategy proof and 4-approximates
the total distance \cite{proportional}. With three or more facilities, they
note that this mechanism is no longer strategy proof. 
With any fixed number $m$ of facilities ($m \geq 1)$ and any 
type of metric space, Fotakis and Tzamos prove that
the winner imposing version of the \proportional\
mechanism is strategy proof and achieves an
approximation ratio for the total distance of at most $4m$
\cite{ft2010}. 
Finally, with $n-1$ facilities and $n$ agents and any
type of metric space, 
Escoffier {\it et al.} prove that the randomized \invprop\ 
mechanism is strategy proof and 
an $n/2$-approximation for the total distance and an $n$-approximation
for the maximum distance \cite{egktps2011}. 
\fi

\section{Conclusions}

We have studied the impact on strategy proof mechanisms for facility location
when moving from one dimension 
to two (or more) dimensions with Euclidean or Manhattan distances. We considered two additional 
axiomatic properties: anonymity which is a fundamental
fairness property, and Pareto optimality which is a fundamental
efficiency property.  We also consider how well such mechanisms
can approximate the optimal welfare. 
Our results are somewhat 
negative. Moving from one dimension to two (or more) dimensions
often makes these axiomatic properties more difficult to achieve. 
For example, with two facilities in Euclidean space
or with just a single facility in Manhattan space, 
no mechanism is anonymous, Pareto optimal
and strategy proof.
By contrast, in one dimension when locating up to two
facilities on the line, there exists mechanisms that are
anonymous, Pareto optimal and strategy proof. 
\ifjournal As a second example, with two facilities 
in Euclidean space
no \percentile\ mechanism has a bounded 
approximation ratio for the total or maximum distance. 
By contrast, when locating up to two
facilities on the line, there exists \percentile\ mechanisms that bound
these approximation ratios. Indeed, such mechanisms can have optimal
approximation ratios. 
\fi
All our impossibility results are {\em minimal}. 
If we drop one of the three axioms (anonymity, Pareto optimality
or strategy proofness) {\em multiple}
mechanisms satisfy the other two axioms. 
For example, when locating two facilities in Euclidean space, 
there exist multiple mechanisms that are anonymous and Pareto optimal, 
that are anonymous and strategy proof, and that are
Pareto optimal and strategy proof.

\eject 
\bibliographystyle{aaai21}
\bibliography{/Users/tw/Documents/biblio/a-z,/Users/tw/Documents/biblio/a-z2,/Users/tw/Documents/biblio/pub,/Users/tw/Documents/biblio/pub2}

\end{document}